\documentclass{article}

\usepackage[nonatbib,final]{neurips_2024}

\usepackage[colorlinks = true, urlcolor  = teal, citecolor=teal]{hyperref}
\usepackage[utf8]{inputenc} 
\usepackage[T1]{fontenc}    
\usepackage{hyperref}       
\usepackage{url}            
\usepackage{booktabs}       
\usepackage{amsfonts}       
\usepackage{nicefrac}       
\usepackage{microtype}      
\usepackage{xcolor}         
\usepackage{amsmath}
\usepackage{graphicx}
\usepackage{listings}
\usepackage[numbers]{natbib}
\usepackage{nicefrac}
\usepackage{rotating}
\usepackage{longtable}
\usepackage{lscape}
\usepackage{wrapfig}
\usepackage{amsmath}
\DeclareMathOperator*{\argmax}{arg\,max}

\title{Evaluating alignment between humans and neural network representations in image-based learning tasks}

\author{%
Can Demircan$^{1, 2, *}$
\quad Tankred Saanum$^2$
\quad \textbf{Leonardo Pettini$^{3, 4}$} 
\quad \textbf{Marcel Binz$^{1, 2}$} \\
\quad \textbf{Blazej M Baczkowski$^{4, 5}$}
\quad \textbf{Christian F Doeller$^{4, 6, 7, 8}$}
\quad \textbf{Mona M Garvert$^{4, 9, 10}$}
\quad \textbf{Eric Schulz$^{1, 2}$} \vspace{3mm} \\
  $^1$Institute for Human-Centered AI, Helmholtz Computational Health Center, Munich, Germany\\
  $^2$Max Planck Institute for Biological Cybernetics - Tübingen, Germany\\
  $^3$Max Planck School of Cognition - Leipzig, Germany\\
  $^4$Max Planck Institute for Human Cognitive \& Brain Sciences - Leipzig, Germany\\
  $^5$University of Tübingen - Tübingen, Germany\\
  $^6$Kavli Institute for Systems Neuroscience - Trondheim, Norway\\
  $^7$Leipzig University - Leipzig, Germany\\
  $^8$Technical University Dresden - Dresden, Germany\\
  $^9$Julius-Maximilians-Universität Würzburg - Würzburg, Germany\\
  $^{10}$Max Planck Institute for Human Development - Berlin, Germany\\
  $^*$\{can.demircan@helmholtz-munich.de\}
}

\begin{document}

\maketitle

\begin{abstract}

Humans represent scenes and objects in rich feature spaces, carrying information that allows us to generalise about category memberships and abstract functions with few examples. What determines whether a neural network model generalises like a human? We tested how well the representations of $86$ pretrained neural network models mapped to human learning trajectories across two tasks where humans had to learn continuous relationships and categories of natural images. In these tasks, both human participants and neural networks successfully identified the relevant stimulus features within a few trials, demonstrating effective generalisation. We found that while training dataset size was a core determinant of alignment with human choices, contrastive training with multi-modal data (text and imagery) was a common feature of currently publicly available models that predicted human generalisation. Intrinsic dimensionality of representations had different effects on alignment for different model types. Lastly, we tested three sets of human-aligned representations and found no consistent improvements in predictive accuracy compared to the baselines. In conclusion, pretrained neural networks can serve to extract representations for cognitive models, as they appear to capture some fundamental aspects of cognition that are transferable across tasks. Both our paradigms and modelling approach offer a novel way to quantify alignment between neural networks and humans and extend cognitive science into more naturalistic domains.
\end{abstract}

\section{Introduction}

Research on representational alignment between neural networks and humans has gained significant attention in recent years \cite{sucholutsky2023getting, schrimpf2018brain}. Comparisons across the systems have provided important insights into neural network representations \cite{muttenthaler_human_2023, sucholutsky2024alignment}, human cognition and the brain \cite{khaligh2014deep, zhuang2021unsupervised, Tuckute2023, schrimpf2021neural, conwell2024large}, and the development of more robust machine learning systems \cite{muttenthaler2024improving, fel2022harmonizing, fu2023dreamsim}.  In the sensory domain, the comparisons have been predominantly made through two families of behavioural tasks. One common approach is to compare object recognition performance across humans and neural networks \cite{rajalingham2018large}. This is a fruitful approach for understanding if the two systems use the same features for object recognition \cite{geirhos2021partial,geirhos2018imagenet, baker2018deep}, are susceptible to similar distortions \cite{geirhos2018generalisation, richardwebster2018psyphy, dodge2017study, hosseini2017limitation}, and struggle with similar images \cite{meding2021trivial}. Another common approach is to use similarity judgement tasks, which may entail reporting pairwise similarity scores \cite{jozwik2017deep, peterson2018evaluating, marjieh2022words}, arranging stimuli in a 2D space based on their similarity \cite{cichy2019spatiotemporal, king2019similarity}, or choosing the odd-one-out in triplets of stimuli \cite{hebart2020revealing, roads2021enriching}. Using these tasks, previous work has identified the factors that contribute to neural networks representing stimuli similarly to humans, both in low-level perception \cite{zhang2018unreasonable} and semantic judgements \cite{muttenthaler_human_2023}.

However, similarity judgements do not begin to capture the complexity of tasks humans use their representations for. Humans rely on rich representations for making judgements and acting in the world. For example, an apple has a multitude of features, such as colour, taste, shape, and brand. Depending on the context, people can use these features and make predictions about the apple's taste, the environmental impacts of growing it, or the significance of it in different mythological and religious settings. What determines whether a neural network model represents an object like an apple with the same richness and flexibility?

In this work, we investigated people's ability to learn functional relationships on naturalistic images in a few-shot setting, and what neural network models best predict human choices. We adapted two commonly used learning paradigms from the cognitive psychology literature: category learning (Fig. \ref{fig:task}A) and reward learning (Fig. \ref{fig:task}B). However, instead of using repeating artificial stimuli, we presented human participants with unique naturalistic images sampled from the THINGS database \cite{hebart_things_2019} in each trial, requiring them to continuously generalise. To understand whether neural networks contain sufficiently rich representations that allow for such generalisation, we tested $86$ different neural networks \cite{hebart2020revealing, dosovitskiy2020image, he2016deep, liu2021swin,liu2022convnet,kubilius2018cornet,fel2022harmonizing, caron2021emerging,misra2020self,caron2020unsupervised,chen2020simple,bardes2021vicreg,zbontar2021barlow,gidaris2018unsupervised,chen2020improved,noroozi2016unsupervised,radford2021learning,cer2018universal,devlin2019bertpretrainingdeepbidirectional,sanh2020distilbertdistilledversionbert,liu2019robertarobustlyoptimizedbert, brown2020languagemodelsfewshotlearners,mikolov2017advancespretrainingdistributedword,muttenthaler2024improving,fu2023dreamsim, mu2021slip, oquab2024dinov2learningrobustvisual, ilharco_gabriel_2021_5143773, Cherti_2023}. These networks varied in their loss function, training diet, and the modality of training data. In summary, we found that:

\begin{figure*}[t]
    \begin{center}
        \includegraphics[width=\textwidth]{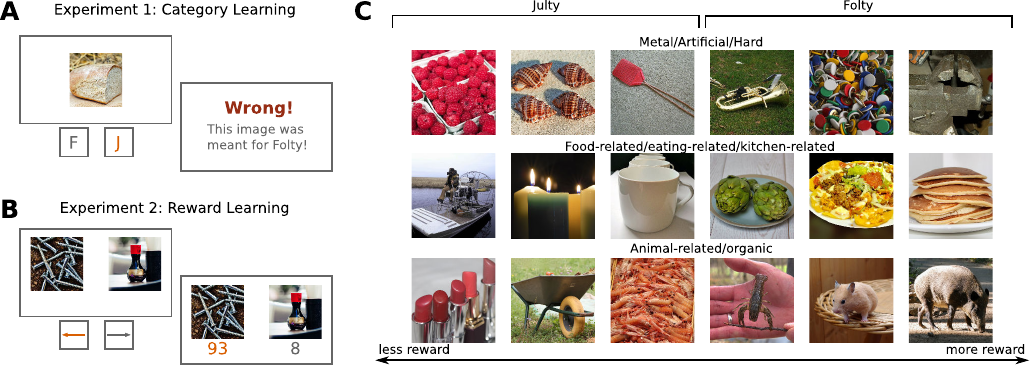}
    \end{center}

    \caption{
    Task descriptions. (\textit{A}) An example trial from the category learning task, where an incorrect decision is made. (\textit{B}) An example trial from the reward learning task where the best option is chosen and highlighted in orange. (\textit{C}) Example images from the THINGS database \cite{hebart_things_2019}. The database has a low dimensional semantically interpretable embedding \cite{hebart2020revealing}, which is derived from human similarity judgements. The example images are placed in the most three prominent dimensions of this embedding. In both tasks, participants were randomly assigned to one of these three dimensions. The associated category membership and rewards for the two tasks are displayed. 
    }
    \label{fig:task}
\end{figure*}

\begin{itemize}
    \item While almost all pretrained models generalised above chance level and predicted human behaviour in both naturalistic learning tasks, contrastive language image pretraining (CLIP) \cite{radford2021learning} consistently yielded the best predictions of human behaviour. We furthermore showed that this could not be fully attributed to the training diet alone.
    \item Multiple factors were important for human alignment, including task performance, model size, training diet, separation of different classes in representations, and the similarity of the representations to the generative embedding of the task.
    \item Of the tested human-aligned neural networks, no method consistently improved human alignment in our tasks compared to non-aligned baselines. However, two of the methods (Harmonization \cite{fel2022harmonizing} and gLocal \cite{muttenthaler2024improving}) yielded improvements in task accuracy on average.
\end{itemize}

\section{Experiments}

We design our experiments around naturalistic images from the THINGS database \cite{hebart_things_2019,hebart2020revealing}. Each image in the database depicts a collection of entities (animals or objects) and comes with an embedding with $49$ human interpretable features, which was built by \citet{hebart2020revealing} to predict human similarity judgements of these objects. Each feature reflects a semantically meaningful property such as whether an image contains metallic objects, food, animals etc. In our experiments, humans learned functions defined over these individual embedding dimensions. We chose category learning and reward learning experiments, as they are well-established paradigms to test function learning and generalisation in human participants. However, unlike traditional paradigms, we used naturalistic images and no images were repeated, requiring generalisation.

\textbf{Category learning:} Human participants ($n=91$) completed $120$ trials of an online category learning task, where they were presented with a novel image in each trial. They were asked to deliver these images to one of two dinosaurs, \textit{Julty} or \textit{Folty}, using key presses. Participants were told that the two dinosaurs had completely non-overlapping preferences for what gifts they enjoyed. After each trial, we gave participants feedback on whether their choice of delivery was correct. An example trial from the task is shown in Fig. \ref{fig:task}\textit{A}. Participants were assigned to one of three conditions, where in each condition the category boundary was defined over a different THINGS embedding dimension. The three chosen dimensions map to how metallic, food-related, and animal-related the shown image is (Fig. \ref{fig:task}\textit{C}). For instance, in one condition non-metallic images should be classified to Folty, and metallic images to Jolty. For each participant, $120$ unique stimuli from the THINGS database were sampled. A median split over the assigned feature of the sampled stimuli determined the category boundary. 

\textbf{Reward learning:} Human participants ($n=82$) completed $60$ trials of a reward learning paradigm \cite{demircan2022decision}, in which they were asked to maximise their accumulated reward throughout the task. In each trial, participants were presented with two images and were asked to select one using key presses. After making a choice, the associated reward with each option was shown. An example trial from the task is shown in Fig. \ref{fig:task}\textit{B}. Participants were assigned to one of the three conditions, as was done in the category learning task. Stimuli were sampled in the same way as the category learning task. For each participant, the values of the task-relevant feature were re-scaled linearly between $0$ and $100$. Additional details about the experimental paradigms are described in Appendix \ref{supp_methods}.

\begin{figure*}[t]
    \begin{center}
    \includegraphics[width=\textwidth]{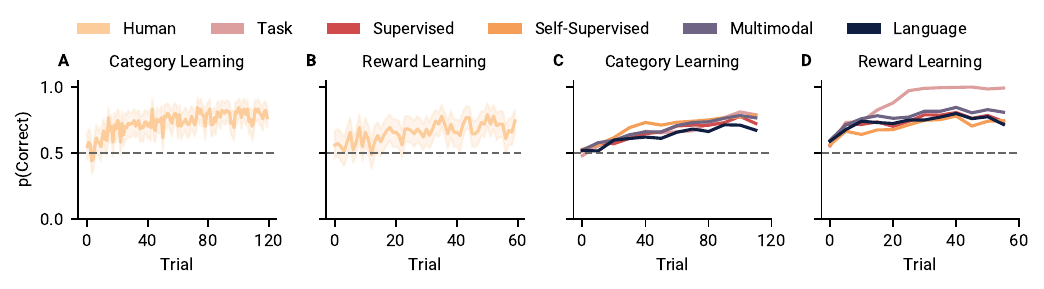}
    \end{center}
    \caption{Learning trajectories of human participants and neural networks. Neural networks can perform as well as humans. (\textit{A} \& \textit{B}) Accuracy of human participants across trials for the category and the reward learning tasks respectively. Shaded lines indicate $95\%$ confidence intervals. (\textit{C} \& \textit{D}) Example learning curves for the neural network representations in the category and the reward learning tasks respectively. The best-performing models from each model type are shown.}
    \label{fig:learning_curves}
\end{figure*}

\section{Behavioural analyses}

\textbf{Humans learn to generalise quickly.} The learning curves of the participants are shown in Fig. \ref{fig:learning_curves}A and \ref{fig:learning_curves}B. To measure whether and how fast people learned in the two experiments, we analysed their choice data using mixed-effects logistic regression models. In the category learning task, we predicted whether a participant made the correct choice using an intercept and the trial number. We found that participants performed this task above chance level, as indicated by a significant intercept ($\hat{\beta} =1.14\pm0.09$, $z=13.18$, $P < .001$),  and that their performance improved over trials ($\hat{\beta} =0.32\pm0.05$, $z=6.89$, $P < .001$), indicating a learning effect. This suggests that people can very efficiently extract the relevant feature dimension in high-dimensional naturalistic environments despite seeing each stimulus only once. For the reward learning task, we predicted whether a participant chose the image on the right using the reward difference between the two images, the trial number, and the interaction of the two predictors. We found that the reward difference ($\hat{\beta} =0.89\pm0.07$, $z=12.56$, $P < .001$), and the interaction of this difference with the trial number ($\hat{\beta} =0.34\pm0.04$, $z=9.30$, $P < .001$) predicted choice, again indicating a learning effect. We further characterise how quickly humans learn the task in Fig. \ref{fig:p_value} in Appendix \ref{add_results} and provide the full specification of the mixed-effects models in Appendix \ref{supp_methods}.

\section{Model-based analyses}

To understand what kind of representations are needed to predict human choices, we tested representations extracted from several pretrained neural networks on our tasks.

\begin{figure*}[t]
\includegraphics[width=\textwidth]{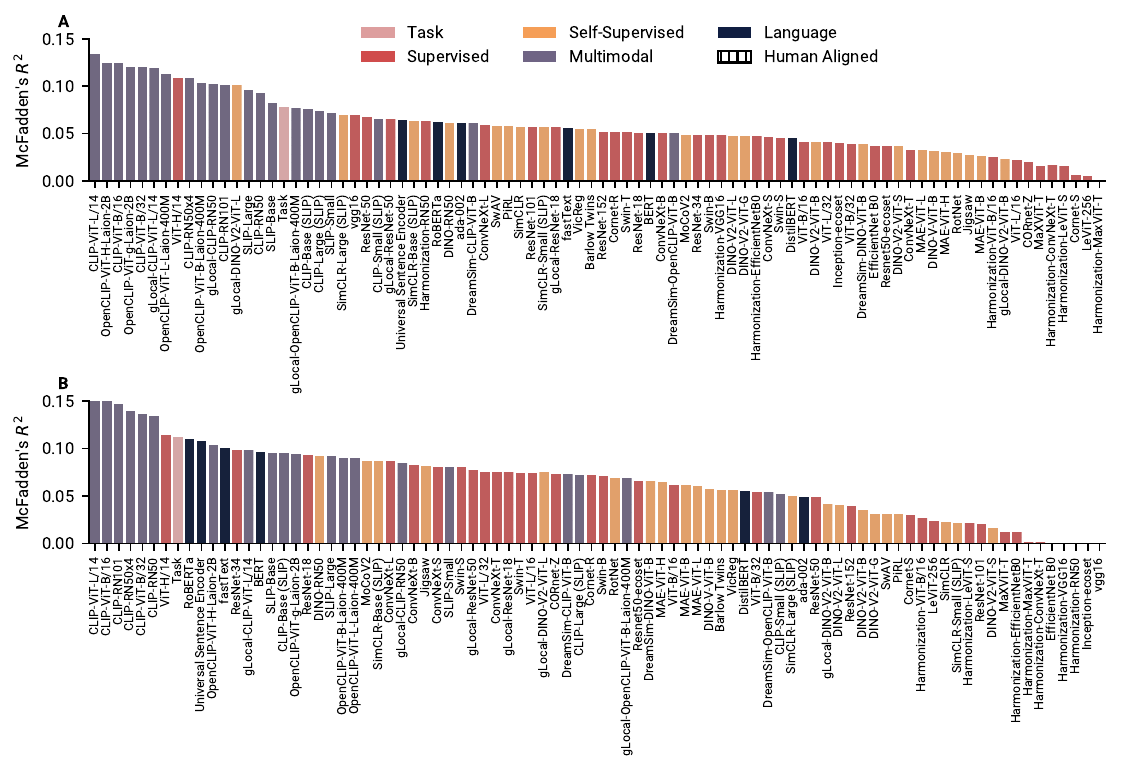}
    \caption{Model fits to human choice data. In both category learning (\textit{A}) and reward learning tasks (\textit{B}), several CLIP models predict human choices the best, even better than the generative features of the tasks. How well the models fitted human choice was more heterogeneously distributed for supervised, self-supervised, and language models. Plotted are the cross-validated McFadden's $R^2$ of each representation for the category learning and the reward learning tasks respectively. Higher values indicate better fits to human behaviour. $0$ marks the alignment of a random model.}
    \label{fig:nll}
\end{figure*}

\textbf{Most representations predict human choice above chance level. CLIP makes the best predictions.} The representations were extracted from the penultimate layer if the models had a classification layer, and from the final layer otherwise. For the transformer models, the \texttt{[CLS]} token representations were extracted. To extract representations from language models, we provided them with the prompt \texttt{A photo of $X$} where \texttt{$X$} was the category label of the task image. fastText was only provided with the category label instead.

We trained linear models to predict either reward or category membership from each neural network model's extracted representations. The models were provided with image-target pairs until trial $t-1$ as training data and made predictions for the image on trial $t$. For the category learning task, we used an  $\ell_2$ regularised logistic regression model, and for the reward learning task, we used a Bayesian linear regression model with spherical Gaussian priors. We used the estimates from the linear models to predict participant choice using mixed-effects logistic regression in leave-one-trial-out cross-validation. For the category learning task, we regressed the probability estimates of the logistic regression models onto participant choice. For the reward learning task, we regressed the reward estimate differences between the left and the right options onto choice. Example learning curves for the two tasks are shown in Fig. \ref{fig:learning_curves}C and Fig. \ref{fig:learning_curves}D. Finally, we measure alignment to human choices using McFadden's $R^2$ \cite{mcfadden1972conditional}, which is computed as follows:

\begin{equation}
    \mathrm{McFadden's} \ R^2 = 1 - \frac{\mathcal{L}_{\text{Model}}}{\mathcal{L}_{\text{Random}}}
\end{equation}

where $\mathcal{L}_{\text{Model}}$ is the negative log likelihood of a given model and $\mathcal{L}_{\text{Random}}$ is the negative log likelihood of a random model.

Most of the representations we tested can do our task and predict human behaviour above chance level across the two tasks (as visualized in Fig. \ref{fig:nll}A and Fig. \ref{fig:nll}B). CLIP models were the top $7$ ($6$) models for the category (reward) learning task in predicting human choices. In total, $16$ ($7$) of the $86$ candidate representations predicted participant behaviour better than the ground truth representations that were used to generate the task. Of these $16$ ($7$) representations,  $14$ ($6$) were CLIP models. One was a large vision transformer, trained in a supervised manner on ImageNet \cite{5206848}. A human-aligned variant of DINO-v2 provided a better fit than the generative task representations in the category learning task. The rest of the supervised and self-supervised vision models, as well as the language models, had a heterogeneous distribution in how well they predicted human behaviour. To provide better intuition for how human participants and CLIP were similar, we display example trials where both CLIP and humans make the same incorrect decisions in Fig. \ref{fig:examples} in Appendix \ref{add_results}.

\subsubsection*{Which factors contribute to alignment?}

\begin{figure*}[t]
    \begin{center}
        \includegraphics[width=\textwidth]{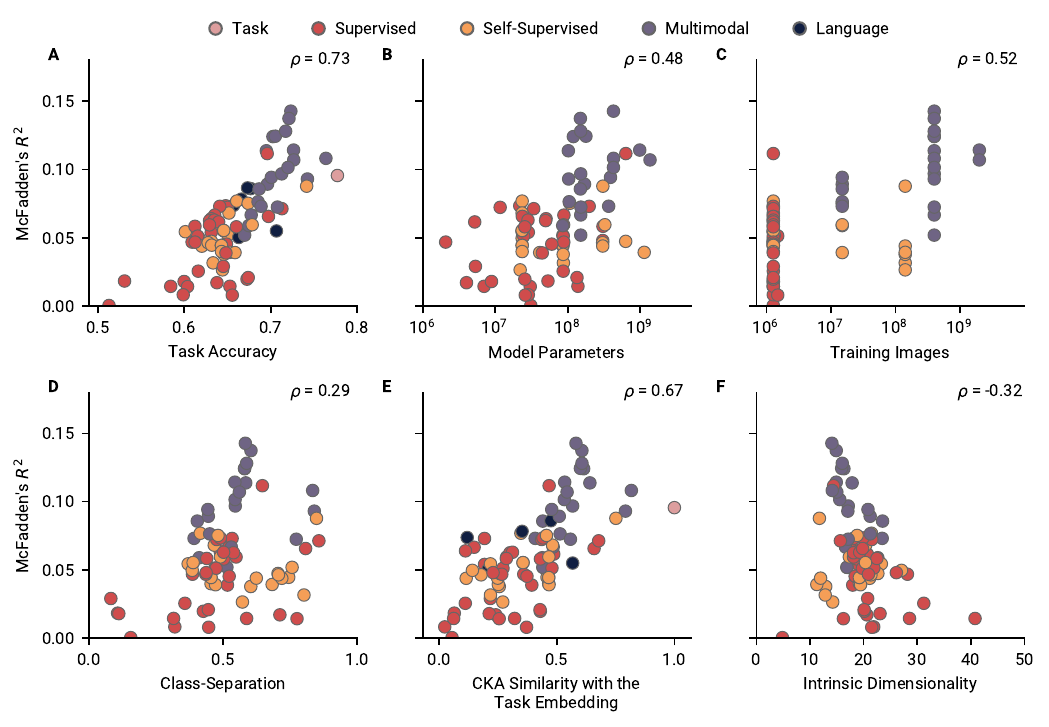}
    \end{center}
    \caption{Several factors contribute to alignment. Models trained on more data and with more trainable parameters predict human choices with higher accuracy. Turning to representations, those that better separate image classes and are more similar to the generative task features exhibit stronger alignment with human choices.}
    \label{fig:rsa}
\end{figure*}

Why are CLIP models substantially better aligned with humans in our task? We conducted a series of analyses to better understand which model properties contribute to alignment. We pooled the data across the two tasks and excluded the language models from all analyses except those shown in Fig. \ref{fig:rsa}A and Fig. \ref{fig:rsa}E, as comparing other properties across vision and language models (e.g. model size) is not meaningful. We first tested if larger models predicted human choice better. While it is common for more expressive models to perform better at downstream computer vision tasks \cite{kolesnikov2020bigtransferbitgeneral, tan2020efficientnetrethinkingmodelscaling, zhai2022scalingvisiontransformers}, previous work has shown that this is not a robust predictor of human alignment \cite{muttenthaler_human_2023, conwell2024large}. In our tasks, we found that larger models predicted human choices better ($\rho = 0.48$, $p < .001$, Fig. \ref{fig:rsa}B), which contradicts previous findings. Next, we considered the number of images seen during training, which is predictive of higher accuracy in image recognition \cite{sun2017revisiting} and human alignment. In our tasks, we found that models trained on more images were more predictive of human choices as well ($\rho = 0.52$, $p < .001$, Fig. \ref{fig:rsa}C).

Then, we analysed which, if any, properties of the models' representations were predictive of their alignment with human choices. First, we considered how well the THINGS classes were separated in the representations of each model. Following \citet{kornblith2021better}, the class-separation was computed as follows:
\begin{equation}
    R^2 = 1 - \bar{d}_\text{within}/\bar{d}_\text{total}\label{eq:class_separation}
\end{equation}
\begin{equation}
    \bar{d}_\text{within} = \sum_{k=1}^K \sum_{m=1}^{N_k} \sum_{n=1}^{N_k} \frac{1 - \mathrm{cos}(\mathbf{x}_{k,m}, \mathbf{x}_{k,n})}{KN_k^2} \quad
    \bar{d}_\text{total} = \sum_{j=1}^K\sum_{k=1}^K\sum_{m=1}^{N_j} \sum_{n=1}^{N_k} \frac{1 - \mathrm{cos}(\mathbf{x}_{j,m}, \mathbf{x}_{k,n})}{K^2 N_j N_k}
\end{equation}
where $\mathbf{x}_{k, m}$ is the representation of image $m$ in object class $k$. $K$ is the total number of classes, and $N_k$ is the total number of images in class $k$. $\mathrm{cos}(\cdot, \cdot)$ denotes cosine similarity between representations. The $R^2$ measure is between 0 and 1, where higher scores indicate a low within-class distance to across-class distance ratio, i.e. high class separation. Previous work has shown a positive link between class separation and image classification \cite{kornblith2021better}, as well as recall \cite{roth2020revisitingtrainingstrategiesgeneralization}. Similar to these findings, we found that models that had higher class separation were more predictive of human choices ($\rho = 0.29$, $p = .01$, Fig. \ref{fig:rsa}D).

We then considered whether the similarity of the representations with the generative task features was predictive of how well different representations predicted human choices. For this, we used linear Centered Kernel Alignment (CKA) \cite{kornblith2019similarityneuralnetworkrepresentations}, which computes the similarity between the generative task representations  $\mathbf{T}$ and neural network representations $\mathbf{X}$ as follows:

\begin{equation} 
\mathrm{CKA}(\mathbf{T},\mathbf{X}) = \dfrac{||\mathbf{X}^T \mathbf{T}||^2_F}{||\mathbf{T}^T\mathbf{T}||_F||\mathbf{X}^T\mathbf{X}||_F} 
\end{equation}
where $||\cdot||_F$ denotes the Frobenius norm. We found that representations that were more similar to the generative task embedding predicted human choices better ($\rho = 0.67$, $p < .001$, Fig. \ref{fig:rsa}E). \begin{wrapfigure}[17]{r}{0.4\textwidth}
  \begin{center}
    \includegraphics[width=0.4\textwidth]{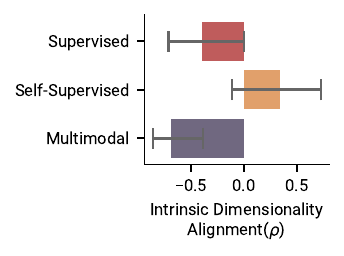}
  \end{center}
  \caption{Lower intrinsic dimensionality is linked with higher alignment most strongly for the multimodal models, and to a lesser extent with supervised ones.}\label{fig:id_factor}
\end{wrapfigure}

Lastly, we tested whether the intrinsic dimensionality of representations was related to alignment. Lower intrinsic dimensionality of neural networks in late layers is positively linked to better classification performance \cite{ansuini2019intrinsic}. The degree to which a network \emph{compresses} its inputs is also directly linked to its ability to generalize \cite{bassily2018learners, tishby2000informationbottleneckmethod, saanum2024reinforcement, saanum2024predicting,shwartz2017opening}. In the human alignment literature, a similar measure named expressed dimensionality has been studied in the context of neural representations. However, diverging from \citet{ansuini2019intrinsic}, one study found a negative correlation between alignment and this measure \cite{elmoznino2024high}, and another study found no link \cite{conwell2024large}. We used the TwoNN method proposed by \citet{facco2017estimating} to estimate intrinsic dimensionality, which makes use of the nearest neighbour distances. First, we linearly scaled all the features to be between $0$ and $1$. We then computed pairwise distances for each pair of data points. Then, we calculated $\mu_i = \nicefrac{r_2}{r_1}$ where $r_1$ and $r_2$ are the shortest distances from datapoint $i$. Later, the empirical cumulative distribution $F^{emp}(\mu)$ was computed by sorting $\mu_i$ and normalising by the total data points $N$. The slope of a linear model that maps $\log \mu_i$ to  $-\log 1 - F^{emp}(\mu_i)$ with no intercept gives the intrinsic dimensionality measure. 

Pooling over all model types, lower intrinsic dimensionality was significantly associated with alignment ($\rho = -0.32$, $p = .03$, Fig. \ref{fig:rsa}F). However, we found that this relationship was most strongly driven by the multimodal models  and to a lesser extent by supervised models (Fig. \ref{fig:id_factor}). That input compression and dataset size are positively related to alignment most strongly for CLIP models suggests that the contrastive multimodal training regime unlocks desirable scaling properties in these models. See  Fig. \ref{fig:pairwise} in Appendix \ref{add_results} for pairwise correlations between the investigated factors.

\subsubsection*{Are CLIP models well aligned only due to their high data diet?}

While we found that models trained with contrastive language image loss predicted human behaviour the best, there remains an important confound. These models are also the ones that are trained on the largest datasets ($400$M to $2$B images). Therefore the direct benefits of multimodal training remain unclear. To address this point, we turned to models provided by \citet{mu2021slip}. Here, the same models are trained on a large dataset (YFCC15M  \cite{Thomee_2016, radford2021learning}) using three different losses: i) a CLIP loss that penalises for the distance between corresponding pairs of text and image representations ii) a SimCLR \cite{chen2020simple} loss that pushes the representations of the augmented and the original image close to each other and away from others, and iii) a CLIP + SimCLR loss. 

\begin{figure*}[t]
    \begin{center}
        \includegraphics[width=\textwidth]{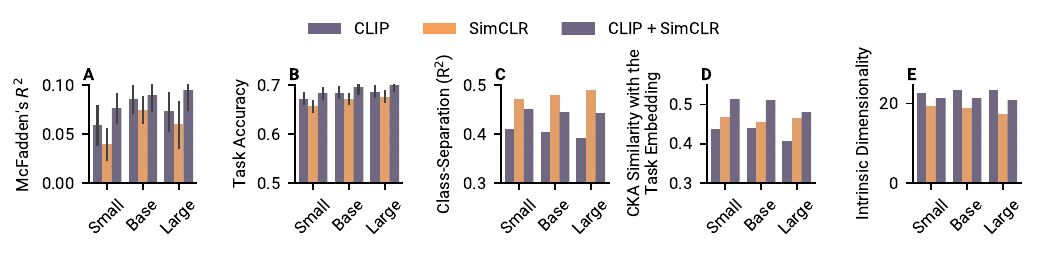}
    \end{center}
    \vspace*{-5mm}

    \caption{The effect of CLIP loss while controlling for model size and data. We observed that CLIP loss increases alignment when data size and architecture are controlled. Here plotted are (\textit{A}) McFadden's $R^2$, (\textit{B}) task accuracy, (\textit{C}) class-separation, (\textit{D}) similarity with the task embedding, and (\textit{E}) intrinsic dimensionality across model sizes and loss functions.}

    \label{fig:slip}
\end{figure*}

First, we found that CLIP models always fit human data better than SimCLR models, and CLIP + SimCLR models made the best predictions when controlling for model size (Fig. \ref{fig:slip}A). This suggests that the advantage provided by the CLIP models cannot solely be attributed to the training data. We found the same ranking of models in terms of how well they did the tasks (Fig. \ref{fig:slip}B). Yet, contrary to our expectations, the SimCLR models had a higher class separation than CLIP models (Fig. \ref{fig:slip}C), as well as better alignment with the generative task features (Fig. \ref{fig:slip}D), and lower intrinsic dimensionality (Fig. \ref{fig:slip}E). This was surprising because, in our previous analyses, we found these properties to be associated with models that predicted human choice better. However, there still may be other confounds that impacted the findings. For example, controlling for training data is not straightforward, as text-image pairs may carry more information than augmented versions of the same image, providing an unfair advantage to the multimodal models.

\subsubsection*{Do alignment methods transfer to our learning tasks?}

Lastly, we evaluated the performance of models that were explicitly aligned to be more human-like. This comparison included three sets of models. \citet{fel2022harmonizing} have aligned models through a method called Harmonization. In addition to the standard supervised training, the models are trained to use the same visual features of images that humans use. The second part is achieved by aligning the networks’ saliency maps with feature importance maps obtained from human judgment. This results in networks that perform better in ImageNet and that are aligned with humans. Next, \citet{fu2023dreamsim} have curated human similarity judgements on a carefully created synthetic dataset. They later fine-tuned pretrained models such as CLIP using Low-Rank Adaptation \cite{hu2021lora} to derive a metric named DreamSim that outperforms other models in predicting human similarity judgements. Lastly,  \citet{muttenthaler2024improving} have fine-tuned representations of pretrained models through a novel transformation named gLocal, which aligns the global representational space to be more human-like by trying to predict human similarity judgements, while preserving the local structure through a contrastive loss that encourages the representations to stay close to their original positions. For these comparisons, we used the models openly provided by the authors.

\begin{figure*}[t]
    \begin{center}
        \includegraphics[width=\textwidth]{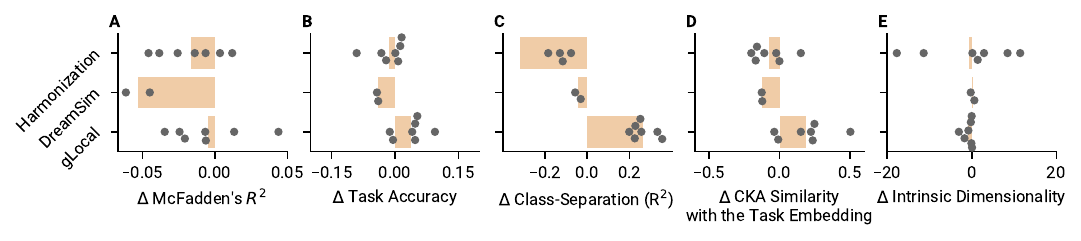}
    \end{center}
    \vspace*{-5mm}
    \caption{We compared models aligned to humans through three different methods against baselines that had the same architecture and that were pretrained on the same data. We found no consistent improvement in human alignment. Here plotted are (\textit{A}) McFadden's $R^2$, (\textit{B}) task accuracy, (\textit{C}) class-separation, (\textit{D}) similarity with the task embedding, and (\textit{D}) intrinsic dimensionality across model sizes and loss functions.}
    \label{fig:alignment}
\end{figure*}

First, we found that none of the alignment methods improved alignment in our task consistently, with some instances of Harmonized and gLocal models improving alignment (Fig. \ref{fig:alignment}A). Alignment improved task accuracy on average for Harmonized and gLocal models (Fig. \ref{fig:alignment}B). Class separation was lower for all Harmonization and DreamSim models, whereas it increased for all gLocal models tested (Fig. \ref{fig:alignment}C). We also observed that the similarity between the representations and the task embedding decreased after alignment for Harmonization and DreamSim, but it increased in most of the models after gLocal alignment (Fig. \ref{fig:alignment}D). Lastly, we observed heterogeneous patterns in the change of intrinsic dimensionality across the three alignment methods, with gLocal reducing the intrinsic dimensionality for all but one of the tested models (Fig. \ref{fig:alignment}E).

\subsection*{How do our tasks compare to other alignment measures?}

\begin{figure*}[t]
    \begin{center}
        \includegraphics[width=\textwidth]{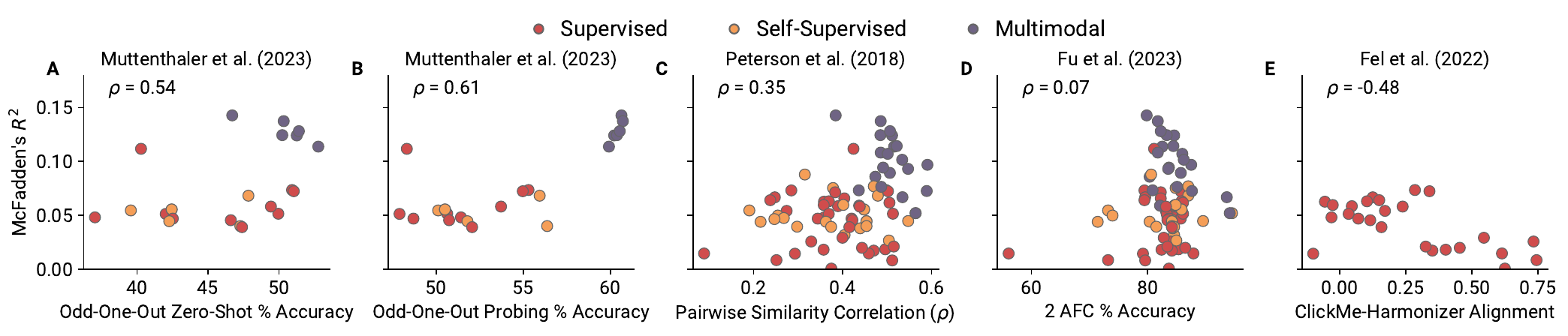}
    \end{center}
    \caption{How do our tasks compare to other alignment methods? Our tasks offer similar (but not identical) results with two of the four similarity judgement tasks. There is a strong negative relationship with the ClickMe dataset, which focuses on localised pixel-level alignment.}
    \label{fig:alignment_tasks}
\end{figure*}

Lastly, to better characterise how our cognitive tasks fit in the alignment literature, we compared them to previously established measures (Fig. \ref{fig:alignment_tasks}). We found the strongest correlation with the THINGS odd-one-out judgements \cite{muttenthaler_human_2023, hebart2020revealing} ($\rho = 0.54$ for zero-shot, and $\rho = 0.61$ for probing). Given the two tasks use the same images, and the ground truth of our tasks was constructed from the odd-one-out judgements, this strong relationship is expected. However, the correlations are still moderate, indicating important differences across tasks. Comparisons with an independent similarity judgement \cite{peterson2018evaluating} task showed a weaker correlation ($\rho = 0.35$), and we found no correlation with a fine-grained two-alternative forced choice task \cite{fu2023dreamsim}. Lastly, we compared alignment in our task to alignment on the ClickMe dataset \cite{linsley2018learning}, which was used to build the Harmonization models \cite{fel2022harmonizing}. We observed a negative correlation ($\rho = -0.48$) here, suggesting that pixel-level alignment and semantically bound global image alignment might be at odds.

\section{Discussion}

In this work, we investigated the alignment of neural network representations to humans. To study this, we measured how well different neural network representations predict human choices in two newly developed learning tasks. Of the $86$ tested representations, all but one predicted human choice above chance level. We furthermore identified several important factors for human alignment, such as large model size, training regime, and low intrinsic dimensionality. These results expand on previous work in both human alignment and cognitive modelling. From an alignment perspective, we considered more challenging tasks compared to previous studies. Previous work has predominantly focused on simple image exposure and similarity judgments. We believe our findings complement this research by addressing unexplored aspects of alignment, which are generalisation and information integration across an extended horizon. From a cognitive modelling perspective, we demonstrated that off-the-shelf pretrained neural networks can serve as representations for cognitive models \cite{battleday_capturing_2020}, which allows to push cognitive models into more naturalistic domains.

\subsection{Related work}

How do our findings compare to previous work on alignment? First, previous research has shown that CLIP representations align well with human representations using brain imaging \cite{muttenthaler_thingsvision_2021, conwell2024large, Wang2023} and similarity judgments \cite{muttenthaler_human_2023}. In line with this, we also found that the contrastive language-image loss improved alignment when controlling for data size and architecture. We furthermore found that training on large datasets generally improved alignment. Yet, it remains unclear whether supervised training on massive datasets alone can achieve high alignment similar to that of CLIP models. For example, \citet{muttenthaler_human_2023} showed that models trained in a supervised manner on the JFT-3B dataset \cite{zhai2022scalingvisiontransformers} can outperform CLIP in predicting human similarity judgments. However, since this is a proprietary dataset, we could not make this comparison. 

We observed some findings that diverge from previous studies using different experimental paradigms. \citet{muttenthaler_human_2023} and  \citet{conwell2024large} found no consistent correlation between a model's number of parameters and its alignment with human similarity judgments and visual cortex activity. In contrast, we found that models with more parameters were more predictive of human choices. Another significant divergence is how intrinsic dimensionality relates to alignment.\citet{elmoznino2024high} found that vision models with higher latent dimensionality better predict visual cortex activity, \citet{conwell2024large} found no correlation. In contrast to this, we observed that lower intrinsic dimensionality led to increased alignment for CLIP models. We hypothesize that both of the observed discrepancies are due to the higher cognitive demands required by our tasks, highlighting the importance of studying alignment in more complex settings. That being said, an alternative explanation for the latter discrepancy could be due to differences in measuring latent dimensionality. Both \citet{elmoznino2024high} and \citet{conwell2024large} use the squared sum of the eigenvalues of principal components divided by the sum of squares of eigenvalues, assuming representations lie on a linear manifold. However, previous work shows that later layers in vision models lie on a curved manifold \cite{ansuini2019intrinsic}. Thus, using principal components might not be the best method for this estimation.

Lastly, we found that a method designed for increasing human alignment, DreamSim\cite{fu2023dreamsim}, actually hurt alignment in our tasks. On the other hand, gLocal \cite{muttenthaler2024improving} and Harmonization \cite{fel_harmonizing_2022} improved both performance and human alignment for some models but not all of them. However, the gLocal transform heavily utilises the THINGS dataset, as it made use of the triplet odd-one-out similarity judgement data \cite{hebart2020revealing}, making it difficult to interpret how well it generalises to other settings. Taken together, these results highlight the importance of studying how well different alignment methods transfer across tasks, as we have done in this work.

\subsection{Limitations}

There are several limitations and extensions of our work that deserve to be highlighted. The main limitation concerns the interactions between factors in the tested neural networks, making it difficult to isolate specific factors. For example, we would like to test the influence of loss function keeping all other factors equal. Ideally, we would train all combinations of architectures, model sizes, loss functions, and datasets, but this is computationally infeasible.  

While we controlled for factors such as training data size and architecture in our comparison of CLIP to other models, there may still be confounding variables we haven't accounted for. For instance, it's not straightforward to compare the information content of image-text pairs used in CLIP training to image-only data used in other models. Text-image pairs might inherently carry more information than single images, potentially giving multimodal models an advantage that's difficult to quantify. This and other subtle differences in training paradigms could influence our results in ways that are challenging to isolate and measure. Lastly, there can also be other families of models that may outperform CLIP models we haven’t considered, such as video models, generative models, or image segmentation models.

We furthermore tested only two experimental paradigms. Future research should explore whether the considered representations predict human behaviour with nonlinear task rules and extend to other paradigms. In particular, one should also consider tasks beyond those generated through the embedding from  \citet{hebart2020revealing}. Previous work has shown that it is possible to automatically generate a large set of text-based category learning problems using large language models \cite{jagadish2024ecologically}. It might be interesting to test whether these methods can be extended to generate tasks involving visual stimuli and use these tasks to test whether our findings generalise to a wider setting.

Finally, we only measured human alignment by looking at behaviour. However, to fully confirm our results, it would also be important to investigate the alignment to brain data. Hence, future work should replicate our experiments in an MRI scanner and compare the representations of neural networks to people's brain activity.

\subsection{Conclusion}

The findings presented in this work have implications both for machine learning and cognitive science. For machine learning, our task and modelling approach offers a new way to measure the human alignment of neural network representations and use this as a metric while building human-aligned neural networks. Alignment at this level can pave the way for artificial systems that can generalise across semantically rich tasks, making them more robust and powerful. For cognitive science, our findings create the opportunity to study several other problems in naturalistic settings by showing that people can do learning tasks with naturalistic stimuli and that pretrained neural networks can be used to extract representations for cognitive models. This could open up the door for a whole new cognitive science that uses naturalistic tasks and environments and thereby increase the validity of the cognitive sciences more generally.

\begin{ack}
This work was funded by the Max Planck Society, the Volkswagen Foundation, as well as the Deutsche Forschungsgemeinschaft (DFG, German Research Foundation) under Germany’s Excellence Strategy–EXC2064/1–390727645.
\end{ack}

\newpage

\bibliographystyle{unsrtnat}

\bibliography{main}

\begin{thebibliography}{89}
\providecommand{\natexlab}[1]{#1}
\providecommand{\url}[1]{\texttt{#1}}
\expandafter\ifx\csname urlstyle\endcsname\relax
  \providecommand{\doi}[1]{doi: #1}\else
  \providecommand{\doi}{doi: \begingroup \urlstyle{rm}\Url}\fi

\bibitem[Sucholutsky et~al.(2023)Sucholutsky, Muttenthaler, Weller, Peng, Bobu, Kim, Love, Grant, Groen, Achterberg, et~al.]{sucholutsky2023getting}
Ilia Sucholutsky, Lukas Muttenthaler, Adrian Weller, Andi Peng, Andreea Bobu, Been Kim, Bradley~C Love, Erin Grant, Iris Groen, Jascha Achterberg, et~al.
\newblock Getting aligned on representational alignment.
\newblock \emph{arXiv preprint arXiv:2310.13018}, 2023.

\bibitem[Schrimpf et~al.(2018)Schrimpf, Kubilius, Hong, Majaj, Rajalingham, Issa, Kar, Bashivan, Prescott-Roy, Geiger, et~al.]{schrimpf2018brain}
Martin Schrimpf, Jonas Kubilius, Ha~Hong, Najib~J Majaj, Rishi Rajalingham, Elias~B Issa, Kohitij Kar, Pouya Bashivan, Jonathan Prescott-Roy, Franziska Geiger, et~al.
\newblock Brain-score: Which artificial neural network for object recognition is most brain-like?
\newblock \emph{BioRxiv}, page 407007, 2018.

\bibitem[Muttenthaler et~al.(2023)Muttenthaler, Dippel, Linhardt, Vandermeulen, and Kornblith]{muttenthaler_human_2023}
Lukas Muttenthaler, Jonas Dippel, Lorenz Linhardt, Robert~A. Vandermeulen, and Simon Kornblith.
\newblock Human alignment of neural network representations.
\newblock In \emph{The Eleventh International Conference on Learning Representations}, 2023.
\newblock URL \url{https://openreview.net/forum?id=ReDQ1OUQR0X}.

\bibitem[Sucholutsky and Griffiths(2024)]{sucholutsky2024alignment}
Ilia Sucholutsky and Tom Griffiths.
\newblock Alignment with human representations supports robust few-shot learning.
\newblock \emph{Advances in Neural Information Processing Systems}, 36, 2024.

\bibitem[Khaligh-Razavi and Kriegeskorte(2014)]{khaligh2014deep}
Seyed-Mahdi Khaligh-Razavi and Nikolaus Kriegeskorte.
\newblock Deep supervised, but not unsupervised, models may explain it cortical representation.
\newblock \emph{PLoS computational biology}, 10\penalty0 (11):\penalty0 e1003915, 2014.

\bibitem[Zhuang et~al.(2021)Zhuang, Yan, Nayebi, Schrimpf, Frank, DiCarlo, and Yamins]{zhuang2021unsupervised}
Chengxu Zhuang, Siming Yan, Aran Nayebi, Martin Schrimpf, Michael~C Frank, James~J DiCarlo, and Daniel~LK Yamins.
\newblock Unsupervised neural network models of the ventral visual stream.
\newblock \emph{Proceedings of the National Academy of Sciences}, 118\penalty0 (3):\penalty0 e2014196118, 2021.

\bibitem[Tuckute et~al.(2023)Tuckute, Feather, Boebinger, and McDermott]{Tuckute2023}
Greta Tuckute, Jenelle Feather, Dana Boebinger, and Josh~H. McDermott.
\newblock Many but not all deep neural network audio models capture brain responses and exhibit correspondence between model stages and brain regions.
\newblock \emph{PLOS Biology}, 21\penalty0 (12):\penalty0 e3002366, December 2023.
\newblock ISSN 1545-7885.
\newblock \doi{10.1371/journal.pbio.3002366}.
\newblock URL \url{http://dx.doi.org/10.1371/journal.pbio.3002366}.

\bibitem[Schrimpf et~al.(2021)Schrimpf, Blank, Tuckute, Kauf, Hosseini, Kanwisher, Tenenbaum, and Fedorenko]{schrimpf2021neural}
Martin Schrimpf, Idan~Asher Blank, Greta Tuckute, Carina Kauf, Eghbal~A Hosseini, Nancy Kanwisher, Joshua~B Tenenbaum, and Evelina Fedorenko.
\newblock The neural architecture of language: Integrative modeling converges on predictive processing.
\newblock \emph{Proceedings of the National Academy of Sciences}, 118\penalty0 (45):\penalty0 e2105646118, 2021.

\bibitem[Conwell et~al.(2024)Conwell, Prince, Kay, Alvarez, and Konkle]{conwell2024large}
Colin Conwell, Jacob~S Prince, Kendrick~N Kay, George~A Alvarez, and Talia Konkle.
\newblock A large-scale examination of inductive biases shaping high-level visual representation in brains and machines.
\newblock \emph{Nature Communications}, 15\penalty0 (1):\penalty0 9383, 2024.

\bibitem[Muttenthaler et~al.(2024)Muttenthaler, Linhardt, Dippel, Vandermeulen, Hermann, Lampinen, and Kornblith]{muttenthaler2024improving}
Lukas Muttenthaler, Lorenz Linhardt, Jonas Dippel, Robert~A Vandermeulen, Katherine Hermann, Andrew Lampinen, and Simon Kornblith.
\newblock Improving neural network representations using human similarity judgments.
\newblock \emph{Advances in Neural Information Processing Systems}, 36, 2024.

\bibitem[Fel et~al.(2022)Fel, Rodriguez~Rodriguez, Linsley, and Serre]{fel2022harmonizing}
Thomas Fel, Ivan~F Rodriguez~Rodriguez, Drew Linsley, and Thomas Serre.
\newblock Harmonizing the object recognition strategies of deep neural networks with humans.
\newblock \emph{Advances in neural information processing systems}, 35:\penalty0 9432--9446, 2022.

\bibitem[Fu et~al.(2023)Fu, Tamir, Sundaram, Chai, Zhang, Dekel, and Isola]{fu2023dreamsim}
Stephanie Fu, Netanel Tamir, Shobhita Sundaram, Lucy Chai, Richard Zhang, Tali Dekel, and Phillip Isola.
\newblock Dreamsim: Learning new dimensions of human visual similarity using synthetic data.
\newblock \emph{arXiv preprint arXiv:2306.09344}, 2023.

\bibitem[Rajalingham et~al.(2018)Rajalingham, Issa, Bashivan, Kar, Schmidt, and DiCarlo]{rajalingham2018large}
Rishi Rajalingham, Elias~B Issa, Pouya Bashivan, Kohitij Kar, Kailyn Schmidt, and James~J DiCarlo.
\newblock Large-scale, high-resolution comparison of the core visual object recognition behavior of humans, monkeys, and state-of-the-art deep artificial neural networks.
\newblock \emph{Journal of Neuroscience}, 38\penalty0 (33):\penalty0 7255--7269, 2018.

\bibitem[Geirhos et~al.(2021)Geirhos, Narayanappa, Mitzkus, Thieringer, Bethge, Wichmann, and Brendel]{geirhos2021partial}
Robert Geirhos, Kantharaju Narayanappa, Benjamin Mitzkus, Tizian Thieringer, Matthias Bethge, Felix~A Wichmann, and Wieland Brendel.
\newblock Partial success in closing the gap between human and machine vision.
\newblock \emph{Advances in Neural Information Processing Systems}, 34:\penalty0 23885--23899, 2021.

\bibitem[Geirhos et~al.(2018{\natexlab{a}})Geirhos, Rubisch, Michaelis, Bethge, Wichmann, and Brendel]{geirhos2018imagenet}
Robert Geirhos, Patricia Rubisch, Claudio Michaelis, Matthias Bethge, Felix~A Wichmann, and Wieland Brendel.
\newblock Imagenet-trained cnns are biased towards texture; increasing shape bias improves accuracy and robustness.
\newblock \emph{arXiv preprint arXiv:1811.12231}, 2018{\natexlab{a}}.

\bibitem[Baker et~al.(2018)Baker, Lu, Erlikhman, and Kellman]{baker2018deep}
Nicholas Baker, Hongjing Lu, Gennady Erlikhman, and Philip~J Kellman.
\newblock Deep convolutional networks do not classify based on global object shape.
\newblock \emph{PLoS computational biology}, 14\penalty0 (12):\penalty0 e1006613, 2018.

\bibitem[Geirhos et~al.(2018{\natexlab{b}})Geirhos, Temme, Rauber, Sch{\"u}tt, Bethge, and Wichmann]{geirhos2018generalisation}
Robert Geirhos, Carlos~RM Temme, Jonas Rauber, Heiko~H Sch{\"u}tt, Matthias Bethge, and Felix~A Wichmann.
\newblock Generalisation in humans and deep neural networks.
\newblock \emph{Advances in neural information processing systems}, 31, 2018{\natexlab{b}}.

\bibitem[RichardWebster et~al.(2018)RichardWebster, Anthony, and Scheirer]{richardwebster2018psyphy}
Brandon RichardWebster, Samuel~E Anthony, and Walter~J Scheirer.
\newblock Psyphy: A psychophysics driven evaluation framework for visual recognition.
\newblock \emph{IEEE transactions on pattern analysis and machine intelligence}, 41\penalty0 (9):\penalty0 2280--2286, 2018.

\bibitem[Dodge and Karam(2017)]{dodge2017study}
Samuel Dodge and Lina Karam.
\newblock A study and comparison of human and deep learning recognition performance under visual distortions.
\newblock In \emph{2017 26th international conference on computer communication and networks (ICCCN)}, pages 1--7. IEEE, 2017.

\bibitem[Hosseini et~al.(2017)Hosseini, Xiao, Jaiswal, and Poovendran]{hosseini2017limitation}
Hossein Hosseini, Baicen Xiao, Mayoore Jaiswal, and Radha Poovendran.
\newblock On the limitation of convolutional neural networks in recognizing negative images.
\newblock In \emph{2017 16th IEEE International Conference on Machine Learning and Applications (ICMLA)}, pages 352--358. IEEE, 2017.

\bibitem[Meding et~al.(2021)Meding, Buschoff, Geirhos, and Wichmann]{meding2021trivial}
Kristof Meding, Luca M~Schulze Buschoff, Robert Geirhos, and Felix~A Wichmann.
\newblock Trivial or impossible--dichotomous data difficulty masks model differences (on imagenet and beyond).
\newblock \emph{arXiv preprint arXiv:2110.05922}, 2021.

\bibitem[Jozwik et~al.(2017)Jozwik, Kriegeskorte, Storrs, and Mur]{jozwik2017deep}
Kamila~M Jozwik, Nikolaus Kriegeskorte, Katherine~R Storrs, and Marieke Mur.
\newblock Deep convolutional neural networks outperform feature-based but not categorical models in explaining object similarity judgments.
\newblock \emph{Frontiers in psychology}, 8:\penalty0 1726, 2017.

\bibitem[Peterson et~al.(2018)Peterson, Abbott, and Griffiths]{peterson2018evaluating}
Joshua~C Peterson, Joshua~T Abbott, and Thomas~L Griffiths.
\newblock Evaluating (and improving) the correspondence between deep neural networks and human representations.
\newblock \emph{Cognitive science}, 42\penalty0 (8):\penalty0 2648--2669, 2018.

\bibitem[Marjieh et~al.(2022)Marjieh, Van~Rijn, Sucholutsky, Sumers, Lee, Griffiths, and Jacoby]{marjieh2022words}
Raja Marjieh, Pol Van~Rijn, Ilia Sucholutsky, Theodore~R Sumers, Harin Lee, Thomas~L Griffiths, and Nori Jacoby.
\newblock Words are all you need? language as an approximation for human similarity judgments.
\newblock \emph{arXiv preprint arXiv:2206.04105}, 2022.

\bibitem[Cichy et~al.(2019)Cichy, Kriegeskorte, Jozwik, van~den Bosch, and Charest]{cichy2019spatiotemporal}
Radoslaw~M Cichy, Nikolaus Kriegeskorte, Kamila~M Jozwik, Jasper~JF van~den Bosch, and Ian Charest.
\newblock The spatiotemporal neural dynamics underlying perceived similarity for real-world objects.
\newblock \emph{Neuroimage}, 194:\penalty0 12--24, 2019.

\bibitem[King et~al.(2019)King, Groen, Steel, Kravitz, and Baker]{king2019similarity}
Marcie~L King, Iris~IA Groen, Adam Steel, Dwight~J Kravitz, and Chris~I Baker.
\newblock Similarity judgments and cortical visual responses reflect different properties of object and scene categories in naturalistic images.
\newblock \emph{NeuroImage}, 197:\penalty0 368--382, 2019.

\bibitem[Hebart et~al.(2020)Hebart, Zheng, Pereira, and Baker]{hebart2020revealing}
Martin~N Hebart, Charles~Y Zheng, Francisco Pereira, and Chris~I Baker.
\newblock Revealing the multidimensional mental representations of natural objects underlying human similarity judgements.
\newblock \emph{Nature human behaviour}, 4\penalty0 (11):\penalty0 1173--1185, 2020.

\bibitem[Roads and Love(2021)]{roads2021enriching}
Brett~D Roads and Bradley~C Love.
\newblock Enriching imagenet with human similarity judgments and psychological embeddings.
\newblock In \emph{Proceedings of the ieee/cvf conference on computer vision and pattern recognition}, pages 3547--3557, 2021.

\bibitem[Zhang et~al.(2018)Zhang, Isola, Efros, Shechtman, and Wang]{zhang2018unreasonable}
Richard Zhang, Phillip Isola, Alexei~A Efros, Eli Shechtman, and Oliver Wang.
\newblock The unreasonable effectiveness of deep features as a perceptual metric.
\newblock In \emph{Proceedings of the IEEE conference on computer vision and pattern recognition}, pages 586--595, 2018.

\bibitem[Hebart et~al.(2019)Hebart, Dickter, Kidder, Kwok, Corriveau, Van~Wicklin, and Baker]{hebart_things_2019}
Martin~N. Hebart, Adam~H. Dickter, Alexis Kidder, Wan~Y. Kwok, Anna Corriveau, Caitlin Van~Wicklin, and Chris~I. Baker.
\newblock Things: A database of 1, 854 object concepts and more than 26, 000 naturalistic object images.
\newblock \emph{PLOS ONE}, 14\penalty0 (10):\penalty0 e0223792, October 2019.
\newblock ISSN 1932-6203.
\newblock \doi{10.1371/journal.pone.0223792}.
\newblock URL \url{http://dx.doi.org/10.1371/journal.pone.0223792}.

\bibitem[Dosovitskiy(2020)]{dosovitskiy2020image}
Alexey Dosovitskiy.
\newblock An image is worth 16x16 words: Transformers for image recognition at scale.
\newblock \emph{arXiv preprint arXiv:2010.11929}, 2020.

\bibitem[He et~al.(2016)He, Zhang, Ren, and Sun]{he2016deep}
Kaiming He, Xiangyu Zhang, Shaoqing Ren, and Jian Sun.
\newblock Deep residual learning for image recognition.
\newblock In \emph{Proceedings of the IEEE conference on computer vision and pattern recognition}, pages 770--778, 2016.

\bibitem[Liu et~al.(2021)Liu, Lin, Cao, Hu, Wei, Zhang, Lin, and Guo]{liu2021swin}
Ze~Liu, Yutong Lin, Yue Cao, Han Hu, Yixuan Wei, Zheng Zhang, Stephen Lin, and Baining Guo.
\newblock Swin transformer: Hierarchical vision transformer using shifted windows.
\newblock In \emph{Proceedings of the IEEE/CVF international conference on computer vision}, pages 10012--10022, 2021.

\bibitem[Liu et~al.(2022)Liu, Mao, Wu, Feichtenhofer, Darrell, and Xie]{liu2022convnet}
Zhuang Liu, Hanzi Mao, Chao-Yuan Wu, Christoph Feichtenhofer, Trevor Darrell, and Saining Xie.
\newblock A convnet for the 2020s.
\newblock In \emph{Proceedings of the IEEE/CVF conference on computer vision and pattern recognition}, pages 11976--11986, 2022.

\bibitem[Kubilius et~al.(2018)Kubilius, Schrimpf, Nayebi, Bear, Yamins, and DiCarlo]{kubilius2018cornet}
Jonas Kubilius, Martin Schrimpf, Aran Nayebi, Daniel Bear, Daniel~LK Yamins, and James~J DiCarlo.
\newblock Cornet: Modeling the neural mechanisms of core object recognition.
\newblock \emph{BioRxiv}, page 408385, 2018.

\bibitem[Caron et~al.(2021)Caron, Touvron, Misra, J{\'e}gou, Mairal, Bojanowski, and Joulin]{caron2021emerging}
Mathilde Caron, Hugo Touvron, Ishan Misra, Herv{\'e} J{\'e}gou, Julien Mairal, Piotr Bojanowski, and Armand Joulin.
\newblock Emerging properties in self-supervised vision transformers.
\newblock In \emph{Proceedings of the IEEE/CVF international conference on computer vision}, pages 9650--9660, 2021.

\bibitem[Misra and Maaten(2020)]{misra2020self}
Ishan Misra and Laurens van~der Maaten.
\newblock Self-supervised learning of pretext-invariant representations.
\newblock In \emph{Proceedings of the IEEE/CVF conference on computer vision and pattern recognition}, pages 6707--6717, 2020.

\bibitem[Caron et~al.(2020)Caron, Misra, Mairal, Goyal, Bojanowski, and Joulin]{caron2020unsupervised}
Mathilde Caron, Ishan Misra, Julien Mairal, Priya Goyal, Piotr Bojanowski, and Armand Joulin.
\newblock Unsupervised learning of visual features by contrasting cluster assignments.
\newblock \emph{Advances in neural information processing systems}, 33:\penalty0 9912--9924, 2020.

\bibitem[Chen et~al.(2020{\natexlab{a}})Chen, Kornblith, Norouzi, and Hinton]{chen2020simple}
Ting Chen, Simon Kornblith, Mohammad Norouzi, and Geoffrey Hinton.
\newblock A simple framework for contrastive learning of visual representations.
\newblock In \emph{International conference on machine learning}, pages 1597--1607. PMLR, 2020{\natexlab{a}}.

\bibitem[Bardes et~al.(2021)Bardes, Ponce, and LeCun]{bardes2021vicreg}
Adrien Bardes, Jean Ponce, and Yann LeCun.
\newblock Vicreg: Variance-invariance-covariance regularization for self-supervised learning.
\newblock \emph{arXiv preprint arXiv:2105.04906}, 2021.

\bibitem[Zbontar et~al.(2021)Zbontar, Jing, Misra, LeCun, and Deny]{zbontar2021barlow}
Jure Zbontar, Li~Jing, Ishan Misra, Yann LeCun, and St{\'e}phane Deny.
\newblock Barlow twins: Self-supervised learning via redundancy reduction.
\newblock In \emph{International conference on machine learning}, pages 12310--12320. PMLR, 2021.

\bibitem[Gidaris et~al.(2018)Gidaris, Singh, and Komodakis]{gidaris2018unsupervised}
Spyros Gidaris, Praveer Singh, and Nikos Komodakis.
\newblock Unsupervised representation learning by predicting image rotations.
\newblock \emph{arXiv preprint arXiv:1803.07728}, 2018.

\bibitem[Chen et~al.(2020{\natexlab{b}})Chen, Fan, Girshick, and He]{chen2020improved}
Xinlei Chen, Haoqi Fan, Ross Girshick, and Kaiming He.
\newblock Improved baselines with momentum contrastive learning.
\newblock \emph{arXiv preprint arXiv:2003.04297}, 2020{\natexlab{b}}.

\bibitem[Noroozi and Favaro(2016)]{noroozi2016unsupervised}
Mehdi Noroozi and Paolo Favaro.
\newblock Unsupervised learning of visual representations by solving jigsaw puzzles.
\newblock In \emph{European conference on computer vision}, pages 69--84. Springer, 2016.

\bibitem[Radford et~al.(2021)Radford, Kim, Hallacy, Ramesh, Goh, Agarwal, Sastry, Askell, Mishkin, Clark, et~al.]{radford2021learning}
Alec Radford, Jong~Wook Kim, Chris Hallacy, Aditya Ramesh, Gabriel Goh, Sandhini Agarwal, Girish Sastry, Amanda Askell, Pamela Mishkin, Jack Clark, et~al.
\newblock Learning transferable visual models from natural language supervision.
\newblock In \emph{International conference on machine learning}, pages 8748--8763. PMLR, 2021.

\bibitem[Cer et~al.(2018)Cer, Yang, Kong, Hua, Limtiaco, John, Constant, Guajardo-Cespedes, Yuan, Tar, et~al.]{cer2018universal}
Daniel Cer, Yinfei Yang, Sheng-yi Kong, Nan Hua, Nicole Limtiaco, Rhomni~St John, Noah Constant, Mario Guajardo-Cespedes, Steve Yuan, Chris Tar, et~al.
\newblock Universal sentence encoder for english.
\newblock In \emph{Proceedings of the 2018 conference on empirical methods in natural language processing: system demonstrations}, pages 169--174, 2018.

\bibitem[Devlin et~al.(2019)Devlin, Chang, Lee, and Toutanova]{devlin2019bertpretrainingdeepbidirectional}
Jacob Devlin, Ming-Wei Chang, Kenton Lee, and Kristina Toutanova.
\newblock Bert: Pre-training of deep bidirectional transformers for language understanding, 2019.
\newblock URL \url{https://arxiv.org/abs/1810.04805}.

\bibitem[Sanh et~al.(2020)Sanh, Debut, Chaumond, and Wolf]{sanh2020distilbertdistilledversionbert}
Victor Sanh, Lysandre Debut, Julien Chaumond, and Thomas Wolf.
\newblock Distilbert, a distilled version of bert: smaller, faster, cheaper and lighter, 2020.
\newblock URL \url{https://arxiv.org/abs/1910.01108}.

\bibitem[Liu et~al.(2019)Liu, Ott, Goyal, Du, Joshi, Chen, Levy, Lewis, Zettlemoyer, and Stoyanov]{liu2019robertarobustlyoptimizedbert}
Yinhan Liu, Myle Ott, Naman Goyal, Jingfei Du, Mandar Joshi, Danqi Chen, Omer Levy, Mike Lewis, Luke Zettlemoyer, and Veselin Stoyanov.
\newblock Roberta: A robustly optimized bert pretraining approach, 2019.
\newblock URL \url{https://arxiv.org/abs/1907.11692}.

\bibitem[Brown et~al.(2020)Brown, Mann, Ryder, Subbiah, Kaplan, Dhariwal, Neelakantan, Shyam, Sastry, Askell, Agarwal, Herbert-Voss, Krueger, Henighan, Child, Ramesh, Ziegler, Wu, Winter, Hesse, Chen, Sigler, Litwin, Gray, Chess, Clark, Berner, McCandlish, Radford, Sutskever, and Amodei]{brown2020languagemodelsfewshotlearners}
Tom~B. Brown, Benjamin Mann, Nick Ryder, Melanie Subbiah, Jared Kaplan, Prafulla Dhariwal, Arvind Neelakantan, Pranav Shyam, Girish Sastry, Amanda Askell, Sandhini Agarwal, Ariel Herbert-Voss, Gretchen Krueger, Tom Henighan, Rewon Child, Aditya Ramesh, Daniel~M. Ziegler, Jeffrey Wu, Clemens Winter, Christopher Hesse, Mark Chen, Eric Sigler, Mateusz Litwin, Scott Gray, Benjamin Chess, Jack Clark, Christopher Berner, Sam McCandlish, Alec Radford, Ilya Sutskever, and Dario Amodei.
\newblock Language models are few-shot learners, 2020.
\newblock URL \url{https://arxiv.org/abs/2005.14165}.

\bibitem[Mikolov et~al.(2017)Mikolov, Grave, Bojanowski, Puhrsch, and Joulin]{mikolov2017advancespretrainingdistributedword}
Tomas Mikolov, Edouard Grave, Piotr Bojanowski, Christian Puhrsch, and Armand Joulin.
\newblock Advances in pre-training distributed word representations, 2017.
\newblock URL \url{https://arxiv.org/abs/1712.09405}.

\bibitem[Mu et~al.(2021)Mu, Kirillov, Wagner, and Xie]{mu2021slip}
Norman Mu, Alexander Kirillov, David Wagner, and Saining Xie.
\newblock Slip: Self-supervision meets language-image pre-training, 2021.

\bibitem[Oquab et~al.(2024)Oquab, Darcet, Moutakanni, Vo, Szafraniec, Khalidov, Fernandez, Haziza, Massa, El-Nouby, Assran, Ballas, Galuba, Howes, Huang, Li, Misra, Rabbat, Sharma, Synnaeve, Xu, Jegou, Mairal, Labatut, Joulin, and Bojanowski]{oquab2024dinov2learningrobustvisual}
Maxime Oquab, Timothée Darcet, Théo Moutakanni, Huy Vo, Marc Szafraniec, Vasil Khalidov, Pierre Fernandez, Daniel Haziza, Francisco Massa, Alaaeldin El-Nouby, Mahmoud Assran, Nicolas Ballas, Wojciech Galuba, Russell Howes, Po-Yao Huang, Shang-Wen Li, Ishan Misra, Michael Rabbat, Vasu Sharma, Gabriel Synnaeve, Hu~Xu, Hervé Jegou, Julien Mairal, Patrick Labatut, Armand Joulin, and Piotr Bojanowski.
\newblock Dinov2: Learning robust visual features without supervision, 2024.
\newblock URL \url{https://arxiv.org/abs/2304.07193}.

\bibitem[Ilharco et~al.(2021)Ilharco, Wortsman, Wightman, Gordon, Carlini, Taori, Dave, Shankar, Namkoong, Miller, Hajishirzi, Farhadi, and Schmidt]{ilharco_gabriel_2021_5143773}
Gabriel Ilharco, Mitchell Wortsman, Ross Wightman, Cade Gordon, Nicholas Carlini, Rohan Taori, Achal Dave, Vaishaal Shankar, Hongseok Namkoong, John Miller, Hannaneh Hajishirzi, Ali Farhadi, and Ludwig Schmidt.
\newblock Openclip, July 2021.
\newblock URL \url{https://doi.org/10.5281/zenodo.5143773}.

\bibitem[Cherti et~al.(2023)Cherti, Beaumont, Wightman, Wortsman, Ilharco, Gordon, Schuhmann, Schmidt, and Jitsev]{Cherti_2023}
Mehdi Cherti, Romain Beaumont, Ross Wightman, Mitchell Wortsman, Gabriel Ilharco, Cade Gordon, Christoph Schuhmann, Ludwig Schmidt, and Jenia Jitsev.
\newblock Reproducible scaling laws for contrastive language-image learning.
\newblock In \emph{2023 IEEE/CVF Conference on Computer Vision and Pattern Recognition (CVPR)}, page 2818–2829. IEEE, June 2023.
\newblock \doi{10.1109/cvpr52729.2023.00276}.
\newblock URL \url{http://dx.doi.org/10.1109/CVPR52729.2023.00276}.

\bibitem[Demircan et~al.(2022)Demircan, Pettini, Saanum, Binz, Baczkowski, Doeller, Garvert, and Schulz]{demircan2022decision}
Can Demircan, Leonardo Pettini, Tankred Saanum, Marcel Binz, Blazej~Metody Baczkowski, Christian Doeller, Mona Garvert, and Eric Schulz.
\newblock Decision-making with naturalistic options.
\newblock In \emph{Proceedings of the Annual Meeting of the Cognitive Science Society}, volume~44, 2022.

\bibitem[McFadden(1972)]{mcfadden1972conditional}
Daniel McFadden.
\newblock Conditional logit analysis of qualitative choice behavior.
\newblock 1972.

\bibitem[Deng et~al.(2009)Deng, Dong, Socher, Li, Li, and Fei-Fei]{5206848}
Jia Deng, Wei Dong, Richard Socher, Li-Jia Li, Kai Li, and Li~Fei-Fei.
\newblock Imagenet: A large-scale hierarchical image database.
\newblock In \emph{2009 IEEE Conference on Computer Vision and Pattern Recognition}, pages 248--255, 2009.
\newblock \doi{10.1109/CVPR.2009.5206848}.

\bibitem[Kolesnikov et~al.(2020)Kolesnikov, Beyer, Zhai, Puigcerver, Yung, Gelly, and Houlsby]{kolesnikov2020bigtransferbitgeneral}
Alexander Kolesnikov, Lucas Beyer, Xiaohua Zhai, Joan Puigcerver, Jessica Yung, Sylvain Gelly, and Neil Houlsby.
\newblock Big transfer (bit): General visual representation learning, 2020.
\newblock URL \url{https://arxiv.org/abs/1912.11370}.

\bibitem[Tan and Le(2020)]{tan2020efficientnetrethinkingmodelscaling}
Mingxing Tan and Quoc~V. Le.
\newblock Efficientnet: Rethinking model scaling for convolutional neural networks, 2020.
\newblock URL \url{https://arxiv.org/abs/1905.11946}.

\bibitem[Zhai et~al.(2022)Zhai, Kolesnikov, Houlsby, and Beyer]{zhai2022scalingvisiontransformers}
Xiaohua Zhai, Alexander Kolesnikov, Neil Houlsby, and Lucas Beyer.
\newblock Scaling vision transformers, 2022.
\newblock URL \url{https://arxiv.org/abs/2106.04560}.

\bibitem[Sun et~al.(2017)Sun, Shrivastava, Singh, and Gupta]{sun2017revisiting}
Chen Sun, Abhinav Shrivastava, Saurabh Singh, and Abhinav Gupta.
\newblock Revisiting unreasonable effectiveness of data in deep learning era, 2017.

\bibitem[Kornblith et~al.(2021)Kornblith, Chen, Lee, and Norouzi]{kornblith2021better}
Simon Kornblith, Ting Chen, Honglak Lee, and Mohammad Norouzi.
\newblock Why do better loss functions lead to less transferable features?, 2021.

\bibitem[Roth et~al.(2020)Roth, Milbich, Sinha, Gupta, Ommer, and Cohen]{roth2020revisitingtrainingstrategiesgeneralization}
Karsten Roth, Timo Milbich, Samarth Sinha, Prateek Gupta, Björn Ommer, and Joseph~Paul Cohen.
\newblock Revisiting training strategies and generalization performance in deep metric learning, 2020.
\newblock URL \url{https://arxiv.org/abs/2002.08473}.

\bibitem[Kornblith et~al.(2019)Kornblith, Norouzi, Lee, and Hinton]{kornblith2019similarityneuralnetworkrepresentations}
Simon Kornblith, Mohammad Norouzi, Honglak Lee, and Geoffrey Hinton.
\newblock Similarity of neural network representations revisited, 2019.
\newblock URL \url{https://arxiv.org/abs/1905.00414}.

\bibitem[Ansuini et~al.(2019)Ansuini, Laio, Macke, and Zoccolan]{ansuini2019intrinsic}
Alessio Ansuini, Alessandro Laio, Jakob~H. Macke, and Davide Zoccolan.
\newblock Intrinsic dimension of data representations in deep neural networks, 2019.

\bibitem[Bassily et~al.(2018)Bassily, Moran, Nachum, Shafer, and Yehudayoff]{bassily2018learners}
Raef Bassily, Shay Moran, Ido Nachum, Jonathan Shafer, and Amir Yehudayoff.
\newblock Learners that use little information.
\newblock In \emph{Algorithmic Learning Theory}, pages 25--55. PMLR, 2018.

\bibitem[Tishby et~al.(2000)Tishby, Pereira, and Bialek]{tishby2000informationbottleneckmethod}
Naftali Tishby, Fernando~C. Pereira, and William Bialek.
\newblock The information bottleneck method, 2000.
\newblock URL \url{https://arxiv.org/abs/physics/0004057}.

\bibitem[Saanum et~al.(2024{\natexlab{a}})Saanum, {\'E}ltet{\H{o}}, Dayan, Binz, and Schulz]{saanum2024reinforcement}
Tankred Saanum, No{\'e}mi {\'E}ltet{\H{o}}, Peter Dayan, Marcel Binz, and Eric Schulz.
\newblock Reinforcement learning with simple sequence priors.
\newblock \emph{Advances in Neural Information Processing Systems}, 36, 2024{\natexlab{a}}.

\bibitem[Saanum et~al.(2024{\natexlab{b}})Saanum, Dayan, and Schulz]{saanum2024predicting}
Tankred Saanum, Peter Dayan, and Eric Schulz.
\newblock Predicting the future with simple world models.
\newblock \emph{arXiv preprint arXiv:2401.17835}, 2024{\natexlab{b}}.

\bibitem[Shwartz-Ziv and Tishby(2017)]{shwartz2017opening}
Ravid Shwartz-Ziv and Naftali Tishby.
\newblock Opening the black box of deep neural networks via information.
\newblock \emph{arXiv preprint arXiv:1703.00810}, 2017.

\bibitem[Elmoznino and Bonner(2024)]{elmoznino2024high}
Eric Elmoznino and Michael~F Bonner.
\newblock High-performing neural network models of visual cortex benefit from high latent dimensionality.
\newblock \emph{PLOS Computational Biology}, 20\penalty0 (1):\penalty0 e1011792, 2024.

\bibitem[Facco et~al.(2017)Facco, d’Errico, Rodriguez, and Laio]{facco2017estimating}
Elena Facco, Maria d’Errico, Alex Rodriguez, and Alessandro Laio.
\newblock Estimating the intrinsic dimension of datasets by a minimal neighborhood information.
\newblock \emph{Scientific reports}, 7\penalty0 (1):\penalty0 12140, 2017.

\bibitem[Thomee et~al.(2016)Thomee, Shamma, Friedland, Elizalde, Ni, Poland, Borth, and Li]{Thomee_2016}
Bart Thomee, David~A. Shamma, Gerald Friedland, Benjamin Elizalde, Karl Ni, Douglas Poland, Damian Borth, and Li-Jia Li.
\newblock Yfcc100m: the new data in multimedia research.
\newblock \emph{Communications of the ACM}, 59\penalty0 (2):\penalty0 64–73, January 2016.
\newblock ISSN 1557-7317.
\newblock \doi{10.1145/2812802}.
\newblock URL \url{http://dx.doi.org/10.1145/2812802}.

\bibitem[Hu et~al.(2021)Hu, Shen, Wallis, Allen-Zhu, Li, Wang, Wang, and Chen]{hu2021lora}
Edward~J. Hu, Yelong Shen, Phillip Wallis, Zeyuan Allen-Zhu, Yuanzhi Li, Shean Wang, Lu~Wang, and Weizhu Chen.
\newblock Lora: Low-rank adaptation of large language models, 2021.

\bibitem[Linsley et~al.(2019)Linsley, Shiebler, Eberhardt, and Serre]{linsley2018learning}
Drew Linsley, Dan Shiebler, Sven Eberhardt, and Thomas Serre.
\newblock Learning what and where to attend.
\newblock \emph{International Conference on Learning Representations (ICLR)}, 2019.

\bibitem[Battleday et~al.(2020)Battleday, Peterson, and Griffiths]{battleday_capturing_2020}
Ruairidh~M. Battleday, Joshua~C. Peterson, and Thomas~L. Griffiths.
\newblock Capturing human categorization of natural images by combining deep networks and cognitive models.
\newblock \emph{Nature Communications}, 11\penalty0 (1), October 2020.
\newblock ISSN 2041-1723.
\newblock \doi{10.1038/s41467-020-18946-z}.
\newblock URL \url{http://dx.doi.org/10.1038/s41467-020-18946-z}.

\bibitem[Muttenthaler and Hebart(2021)]{muttenthaler_thingsvision_2021}
Lukas Muttenthaler and Martin~N. Hebart.
\newblock Thingsvision: A python toolbox for streamlining the extraction of activations from deep neural networks.
\newblock \emph{Frontiers in Neuroinformatics}, 15, 2021.
\newblock ISSN 1662-5196.
\newblock \doi{10.3389/fninf.2021.679838}.
\newblock URL \url{https://www.frontiersin.org/journals/neuroinformatics/articles/10.3389/fninf.2021.679838}.

\bibitem[Wang et~al.(2023)Wang, Kay, Naselaris, Tarr, and Wehbe]{Wang2023}
Aria~Y. Wang, Kendrick Kay, Thomas Naselaris, Michael~J. Tarr, and Leila Wehbe.
\newblock Better models of human high-level visual cortex emerge from natural language supervision with a large and diverse dataset.
\newblock \emph{Nature Machine Intelligence}, 5\penalty0 (12):\penalty0 1415–1426, November 2023.
\newblock ISSN 2522-5839.
\newblock \doi{10.1038/s42256-023-00753-y}.
\newblock URL \url{http://dx.doi.org/10.1038/s42256-023-00753-y}.

\bibitem[Jagadish et~al.(2024)Jagadish, Coda-Forno, Thalmann, Schulz, and Binz]{jagadish2024ecologically}
Akshay~K Jagadish, Julian Coda-Forno, Mirko Thalmann, Eric Schulz, and Marcel Binz.
\newblock Ecologically rational meta-learned inference explains human category learning.
\newblock \emph{arXiv preprint arXiv:2402.01821}, 2024.

\bibitem[de~Leeuw(2014)]{deLeeuw2014}
Joshua~R. de~Leeuw.
\newblock jspsych: A javascript library for creating behavioral experiments in a web browser.
\newblock \emph{Behavior Research Methods}, 47\penalty0 (1):\penalty0 1–12, March 2014.
\newblock ISSN 1554-3528.
\newblock \doi{10.3758/s13428-014-0458-y}.
\newblock URL \url{http://dx.doi.org/10.3758/s13428-014-0458-y}.

\bibitem[{R Core Team}(2021)]{rcitation}
{R Core Team}.
\newblock \emph{R: A Language and Environment for Statistical Computing}.
\newblock R Foundation for Statistical Computing, Vienna, Austria, 2021.
\newblock URL \url{https://www.R-project.org/}.

\bibitem[Bates et~al.(2015)Bates, M\"{a}chler, Bolker, and Walker]{Bates2015}
Douglas Bates, Martin M\"{a}chler, Ben Bolker, and Steve Walker.
\newblock Fitting linear mixed-effects models usinglme4.
\newblock \emph{Journal of Statistical Software}, 67\penalty0 (1), 2015.
\newblock ISSN 1548-7660.
\newblock \doi{10.18637/jss.v067.i01}.
\newblock URL \url{http://dx.doi.org/10.18637/jss.v067.i01}.

\bibitem[Pedregosa et~al.(2018)Pedregosa, Varoquaux, Gramfort, Michel, Thirion, Grisel, Blondel, Müller, Nothman, Louppe, Prettenhofer, Weiss, Dubourg, Vanderplas, Passos, Cournapeau, Brucher, Perrot, and Édouard Duchesnay]{pedregosa2018scikitlearnmachinelearningpython}
Fabian Pedregosa, Gaël Varoquaux, Alexandre Gramfort, Vincent Michel, Bertrand Thirion, Olivier Grisel, Mathieu Blondel, Andreas Müller, Joel Nothman, Gilles Louppe, Peter Prettenhofer, Ron Weiss, Vincent Dubourg, Jake Vanderplas, Alexandre Passos, David Cournapeau, Matthieu Brucher, Matthieu Perrot, and Édouard Duchesnay.
\newblock Scikit-learn: Machine learning in python, 2018.
\newblock URL \url{https://arxiv.org/abs/1201.0490}.

\bibitem[Speekenbrink et~al.(2008)Speekenbrink, Channon, and Shanks]{speekenbrink2008learning}
Maarten Speekenbrink, Shelley Channon, and David~R Shanks.
\newblock Learning strategies in amnesia.
\newblock \emph{Neuroscience \& Biobehavioral Reviews}, 32\penalty0 (2):\penalty0 292--310, 2008.

\bibitem[Gershman(2015)]{gershman2015unifying}
Samuel~J Gershman.
\newblock A unifying probabilistic view of associative learning.
\newblock \emph{PLoS computational biology}, 11\penalty0 (11):\penalty0 e1004567, 2015.

\bibitem[Binz et~al.(2022)Binz, Gershman, Schulz, and Endres]{binz_heuristics_2022}
Marcel Binz, Samuel~J. Gershman, Eric Schulz, and Dominik Endres.
\newblock Heuristics from bounded meta-learned inference.
\newblock \emph{Psychological Review}, 129\penalty0 (5):\penalty0 1042–1077, October 2022.
\newblock ISSN 0033-295X.
\newblock \doi{10.1037/rev0000330}.
\newblock URL \url{http://dx.doi.org/10.1037/rev0000330}.

\bibitem[Marjieh(2024)]{Marjieh_2024}
Raja Marjieh.
\newblock The universal law of generalization holds for naturalistic stimuli, Jan 2024.
\newblock URL \url{osf.io/rbkgh}.

\bibitem[Stoinski et~al.(2022)Stoinski, Perkuhn, and Hebart]{stoinski_thingsplus_2022}
Laura~Mai Stoinski, Jonas Perkuhn, and Martin~N Hebart.
\newblock Thingsplus: New norms and metadata for the things database of 1, 854 object concepts and 26, 107 natural object images, July 2022.
\newblock URL \url{http://dx.doi.org/10.31234/osf.io/exu9f}.

\end{thebibliography}


\newpage

\appendix

\section{Methods} \label{supp_methods}

\textbf{Participants} For the category learning task, we recruited $98$ participants ($48$ females, $50$ males, mean age$=28.92$y, SD$=7.32$) on the Prolific platform. Participants with less than $50\%$ accuracy were excluded from the analyses, leaving us with $91$ participants. A base payment of £ $1.50$ was made, and participants could earn an additional bonus of  £ $6.00$. The median completion time was $12$ minutes and $38$ seconds. The inclusion criteria included having a minimum approval rate of $97\%$, and a minimum number of $15$ previous submissions on Prolific. Participation in the reward learning study was an exclusion criterion. For the reward learning task, $99$ participants were recruited ($49$ females, $49$ males, $1$ other, mean age $= 27.9$ y, SD $=9.13$).  After applying the $50\%$ accuracy criteria, we were left with $82$ participants. A base payment of £ $2.00$ was made, and an additional performance-dependent bonus of £$4.00$ was offered. The median completion time was $9$ minutes and $26$ seconds. The inclusion criteria included having a minimum approval rate of $95\%$, and a minimum number of $10$ previous submissions on Prolific. All participants agreed to their anonymized data being used for research. The study was approved by the ethics committee of of the medical faculty of the University of Tübingen (number 701/2020BO). Participants gave consent for their data to be anonymously analyzed by agreeing to a data protection sheet approved by the data protection officer of the MPG (Datenschutzbeauftragte der MPG, Max-Planck-Gesellschaft zur Förderung der Wissenschaften).

\textbf{Tasks and Stimuli} Both tasks were run online in forced full-screen mode. Participants were shown written instructions and were asked to complete comprehension check questions before they could start the tasks. In both tasks, participants were given unlimited time to make decisions. In the category learning task, binary (correct versus wrong) feedback was given for $2s$. In the reward learning task, the associated reward with the stimuli was shown for $1.5s$, and there was an inter-trial interval of $1s$ where participants were shown a blank screen. Throughout both tasks, the estimated total payment of participants was shown on the upper part of the screen. At the end of the tasks, participants were asked whether they thought their data should be used for analysis. Across both tasks, all but one participant responded saying their data should be analyzed, whose data was anyway excluded due to poor performance. The category learning task was programmed using jsPsych \cite{deLeeuw2014}, whereas the reward learning task was programmed in plain JavaScript.

For each participant, $120$ stimuli were sampled independently from the THINGS database. Because the loadings of the features were not uniformly distributed, we made $5$ equally sized bins of the loadings for the assigned feature and sampled object categories uniformly from these bins. From these object categories, the specific images were assigned randomly. For details on the used features and the embedding, see \citet{hebart2020revealing}.

\textbf{Behavioural Analyses} We used mixed-effects logistic models for both category and reward learning analyses. For category learning, we predicted correct responses per trial, using trial number as a fixed effect and including participant-specific random effects for intercept, trial number, and assigned task rule. In the reward learning model, we predicted whether the image on the right is selected, incorporating the trial number, reward difference between images, and their interaction as fixed effects. These factors, along with the assigned task rule, were also modelled as participant-specific random effects. Both models effectively captured task structure, learning progression, and individual variability in performance. In R formula notation\cite{rcitation,Bates2015}, the model for category learning is denoted as follows:

\begin{lstlisting}
correct_choice ~ 1 + trial + (1 + trial + dimension | participant)

\end{lstlisting}

For the reward learning task, the following model was used:

\begin{lstlisting}
right_choice ~ -1 + trial * right_left_reward_difference + 
(-1 + trial + dimension + right_left_reward_difference | participant)
\end{lstlisting}

where $-1$ denotes no intercept.

\textbf{Software, Data, \& Compute Resources} The code to reproduce the reported results is available at \url{https://github.com/candemircan/naturalcogsci} and we provide anonymised human choice data on \url{https://osf.io/h3t52/}. For the learning models we used \texttt{lme4} \cite{Bates2015} and \texttt{scikit-learn} \cite{pedregosa2018scikitlearnmachinelearningpython}. To extract representations from neural networks we used \texttt{thingsvision} \cite{muttenthaler_thingsvision_2021}.

Computations were performed on an academic SLURM cluster. Feature extraction was done on a single Nvidia A100 GPU (40GB) under 24 hours. The linear models were parallelised across several jobs that used single core and 8GB RAM and were completed in under 48 hours. The mixed-effects models were similarly parallelised and completed under 24 hours.

\textbf{Modelling} 

For the category learning task, we used an  $\ell_2$ regularised logistic regression model to optimize regression weights. We relied on \texttt{scikit-learn}’s \texttt{LogisticRegression} class which internally optimizes the following objective:
\begin{equation}
\mathbf{w}^* = \argmax_{\mathbf{w}} \sum_{i=1}^N -c_i \log(p(c_i | \mathbf{x}_i, \mathbf{w})) - (1-c_i) \log(1-\log(p(c_i | \mathbf{x}_i, \mathbf{w})) + \dfrac{1}{2} ||\mathbf{w} ||^2_2 
\end{equation}

For the reward learning task, we used a Bayesian linear regression model to infer a posterior distribution over regression weights. We relied on \texttt{scikit-learn}’s \texttt{BayesianRidge} class which infers a posterior distribution assuming spherical Gaussian priors (i.e., $p(\mathbf{w}) = \mathcal{N}(0, \lambda^{-1}\mathbf{I})$) and Gaussian likelihood (i.e., $p(y_i | \mathbf{x}_i, \mathbf{w}) = \mathcal{N}(\mathbf{w}^{\top}\mathbf{x}_i, \beta^{-1})$). Based on these assumptions, the posterior distribution can be computed in closed form:
\begin{align}
p(\mathbf{w} | \mathbf{X}, \mathbf{y}) &= \mathcal{N}(\mathbf{m}_N, \mathbf{S}_N) \\
\mathbf{m}_N &= \beta \mathbf{S}_N \mathbf{X}^{\top} \mathbf{y} \\
\mathbf{S}^{-1}_N &=  \lambda \mathbf{I} + \beta \mathbf{X}^{\top}\mathbf{X}
\end{align}

where $\mathbf{X}$ and $\mathbf{y}$ denote the stacked inputs and targets respectively.

We run both models from scratch on each trial using all previously observed input-target pairs. The choice of these models was motivated by previous investigations in similar -- but low-dimensional -- settings \cite{speekenbrink2008learning, gershman2015unifying, binz_heuristics_2022}.

The $\ell_2$ penalty term for the logistic regression model described above was determined via grid search to maximise task performance, on a per participant basis. For the linear regression model, $\lambda$ and $\beta$ were fitted to maximise the log marginal likelihood on the task performances. 

The estimates from these models were used in mixed-effects logistic regression models with leave-one-out predictions to assess participant choices. For category learning, we used logistic regression probability estimates as predictors. In the reward learning task, we used the difference in estimated rewards from linear regression models as predictors. In both cases, these predictors were included as both fixed and random effects, allowing us to account for individual differences while maintaining the group effects. These correspond to the following models in R formula notation for category and reward learning respectively:

\begin{lstlisting}
human_choice ~ -1 + probability_estimate + (-1 + probability_estimate | participant)
\end{lstlisting}

\begin{lstlisting}
human_choice ~ -1 + estimated_reward_difference + (-1 + estimated_reward_difference | participant)
\end{lstlisting}

For the mixed-effects models, the training data was centred and divided by its standard deviation. The same scaling parameters were applied to the test data. 

\textbf{Additional human-alignment tasks}

Results reported in Fig. \ref{fig:alignment_tasks}A \& Fig. \ref{fig:alignment_tasks}B include comparisons for the overlap of models between those reported by Muttenthaler et al. \cite{muttenthaler_human_2023} and the ones we tested. For Fig. \ref{fig:alignment_tasks}B \& Fig. \ref{fig:alignment_tasks}C, we tested all the vision models reported in our paper. However, for the Peterson et al. \cite{peterson2018evaluating} dataset, we only found a subset of the original data reported in the paper\footnote{Specifically, we tested similarity judgements obtained on Animal, Fruit, and Vegetable categories. The data was obtained from \cite{Marjieh_2024}}. The ClickMe-Harmonizer alignment was only computed for supervised models, as the method requires computing gradients for ImageNet classes, which we could only do for the supervised models that had ImageNet classification heads.

\section{Testing aligned models on other datasets}

\begin{figure*}[h]
    \begin{center}
        \includegraphics[width=\textwidth]{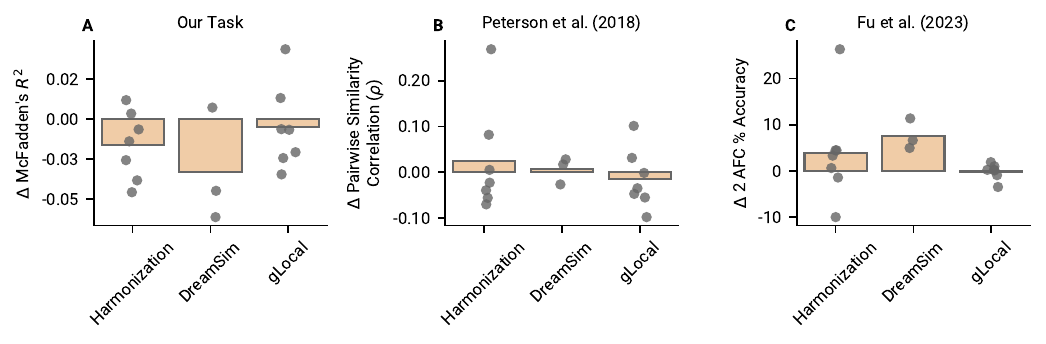}
    \end{center}
    \caption{Change of human alignment for different methods on different datasets}
    \label{fig:alignment_tests}
\end{figure*}

Above, we also report how different alignment methods perform on different datasets(Fig. \ref{fig:alignment_tests}). Harmonization is on average slightly more human-like on the two external tested datasets compared to baselines. DreamSim shows mixed results for the Peterson et al.  \cite{peterson2018evaluating} dataset, but it shows improvement on the NIGHTS dataset \cite{fu2023dreamsim}. This is not surprising, as this dataset was used to build DreamSim. Lastly, gLocal shows mixed results.

\section{Additional results} \label{add_results}

\begin{figure*}[h]
    \begin{center}
        \includegraphics[width=\textwidth]{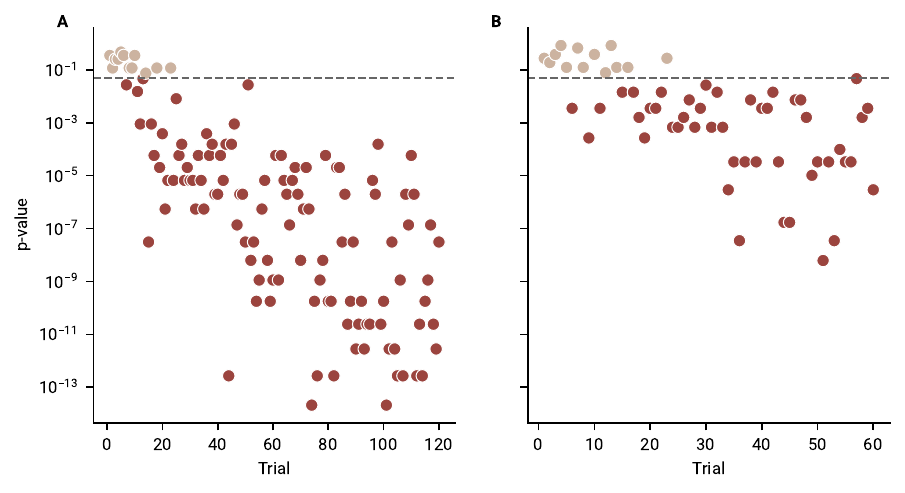}
    \end{center}
    \caption{Participant Performance Against Chance Level at Each Trial. Trial-by-trial p-values from 1 sample t-tests testing accuracy against chance level for (\textit{A}) category learning task and the (\textit{B}) reward learning task.}
    \label{fig:p_value}
\end{figure*}

\begin{figure*}
    \begin{center}
        \includegraphics[width=\textwidth]{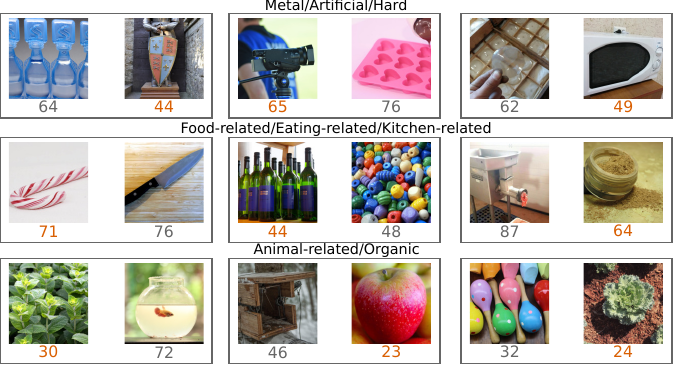}
    \end{center}
    \caption{Example trials showing the similarity between CLIP and human decisions that show disagreement with the task embedding. Each row shows three trials from a different condition. Orange highlighted text shows the option chosen by all CLIP models and the human participant, whereas grey text shows the decision made by the task embedding. As the tasks were generated using the task embedding, all the choices shown here made by CLIP and humans are suboptimal. Shown examples are from the second half of the task, as to eliminate the learning process as a confound. The original images are replaced with copyright-free alternatives from the THINGSplus database \cite{stoinski_thingsplus_2022}.}
    \label{fig:examples}
\end{figure*}

\begin{figure}
    \begin{center}
        \includegraphics[width=\textwidth]{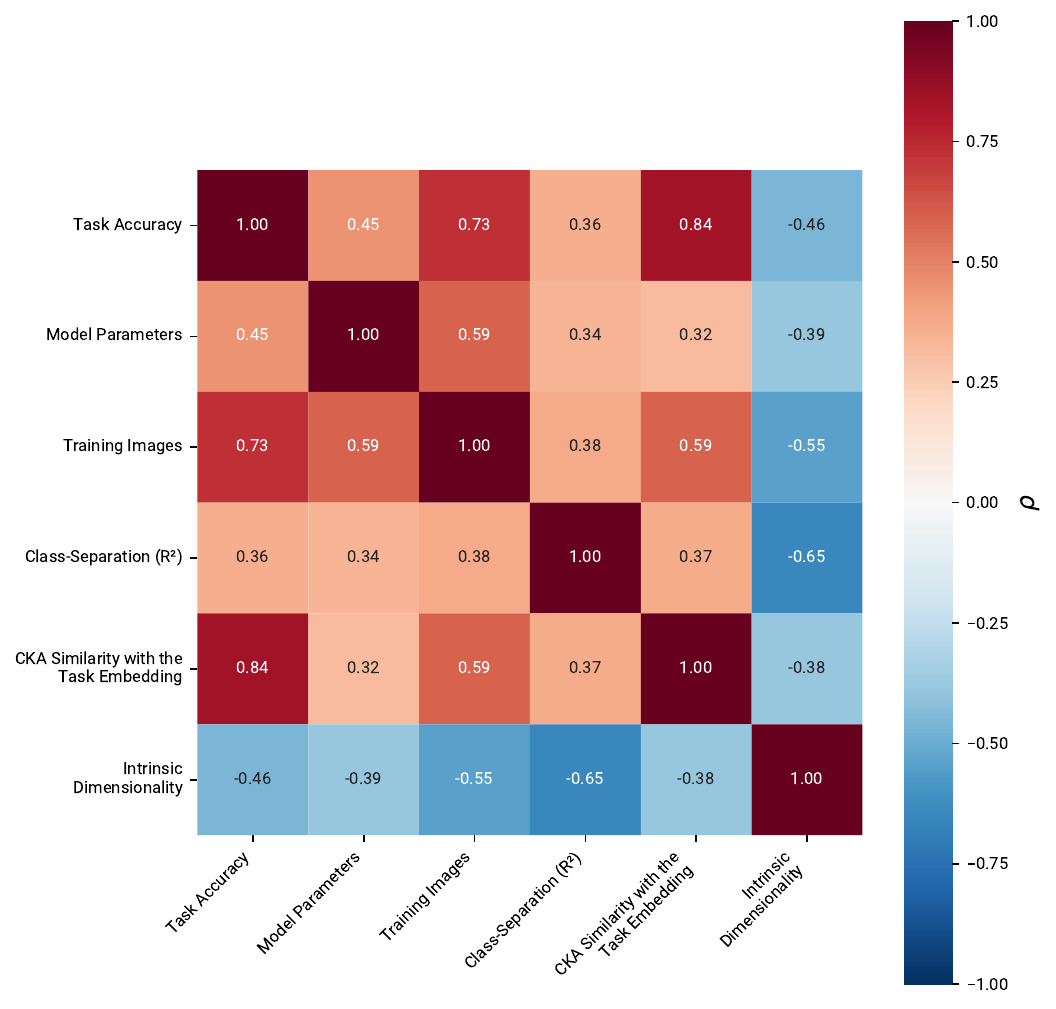}
    \end{center}
    \caption{Pairwise Spearman correlations between the factors investigated that contribute to alignment. }
    \label{fig:pairwise}
\end{figure}

Below we provide some additional results supporting our claims in the main text. Fig. \ref{fig:p_value} shows participants can do both tasks above chance level very early on in the task. Fig. \ref{fig:examples} shows some incorrect choices made by humans and also by CLIP models, and Fig. \ref{fig:pairwise} shows pairwise correlations between the factors we investigated that contribute to alignment.

\end{document}